\documentclass[letterpaper]{article} % DO NOT CHANGE THIS
\usepackage{aaai2027}  % DO NOT CHANGE THIS
\usepackage[hyphens]{url}  % DO NOT CHANGE THIS
\usepackage{graphicx} % DO NOT CHANGE THIS
\usepackage{natbib}  % DO NOT CHANGE THIS AND DO NOT ADD ANY OPTIONS TO IT
\usepackage{caption} % DO NOT CHANGE THIS AND DO NOT ADD ANY OPTIONS TO IT
\usepackage{algorithm}
\usepackage{algorithmic}
\newcommand{\evalunlearn}{\texttt{eval-unlearn}}
\nocopyright
\usepackage{newfloat}
\usepackage{listings}
\DeclareCaptionStyle{ruled}{labelfont=normalfont,labelsep=colon,strut=off} % DO NOT CHANGE THIS
\floatstyle{ruled}
\newfloat{listing}{tb}{lst}{}
\floatname{listing}{Listing}

\usepackage{booktabs}

\title{eval-unlearn: Benchmarking unlearning in Text-to-Image Diffusion Models}
\author {
    Mansi\textsuperscript{\rm 1}\corresponding,
    Nikhil Raghavan\textsuperscript{\rm 1},
    Zixia Huang\textsuperscript{\rm 1},
    Kai Sheng Ong\textsuperscript{\rm 1},
    Ji Shen Lim\textsuperscript{\rm 1},
    Brandon Siao Xiang Ling\textsuperscript{\rm 1},
    Francesco Leofante\textsuperscript{\rm 1}
}
\affiliations {
    \textsuperscript{\rm 1}Imperial College London\\
    \{m.-24, nikhil.raghavan, zixia.huang, kai-sheng.ong,
    dylan.lim, brandon.ling, f.leofante\}@imperial.ac.uk
}
\begin{document}

\maketitle

% \begin{abstract}
% AAAI creates proceedings, working notes, and technical reports directly from electronic source furnished by the authors. To ensure that all papers in the publication have a uniform appearance, authors must adhere to the following instructions.
% \end{abstract}

\begin{figure*}[ht]
    \centering
    \includegraphics[width=1\linewidth]{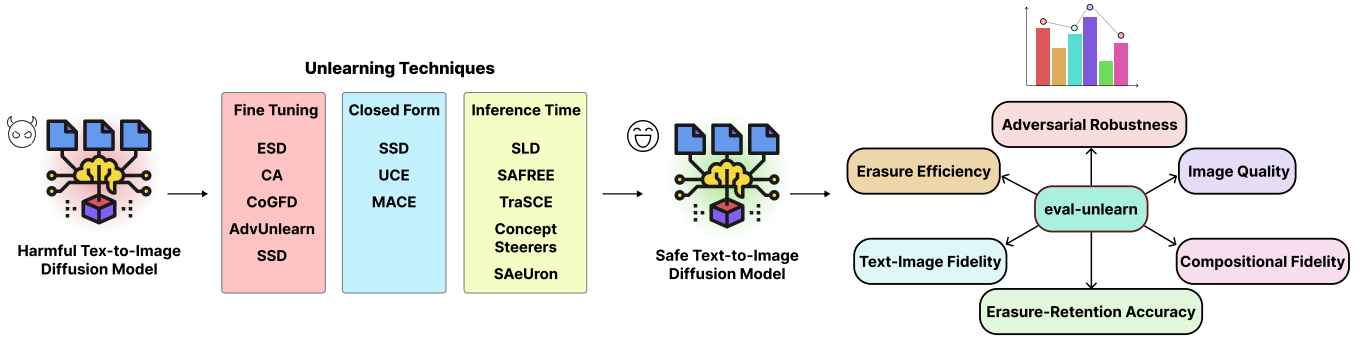}
    \caption{The figure shows the pipeline for \evalunlearn{} on an abstract level.}
    \label{fig:pipeline}
\end{figure*}
\begin{abstract}%

The rising number of concept unlearning techniques for text-to-image (T2I) diffusion models has
produced a fragmented evaluation landscape. Methods are assessed under heterogeneous experimental
conditions making principled cross-method comparison difficult. We present \evalunlearn, an open-source Python library providing a unified,
reproducible benchmarking framework for concept unlearning in T2I Diffusion models. \evalunlearn{}
integrates twelve published unlearning techniques spanning fine-tuning, closed-form model editing,
and inference-time intervention, alongside nine complementary evaluation metrics covering erasure
efficacy, adversarial robustness, generative quality, and concept retention. Its plugin architecture lets third-party techniques and metrics self-register without modifying the core
framework, and its streaming, batched pipeline supports efficient evaluation of both standard NSFW
concepts and arbitrary general concepts. As a further contribution, we release a public
leaderboard on HuggingFace along with an interactive tool for real-time evaluation of unlearning techniques. The leaderboard compares nudity concept
erasure case study across all twelve techniques, exposing significant accuracy-quality trade-offs
that are obscured by heterogeneous evaluation.
\evalunlearn{} is released under the MIT license; the package, code, leaderboard, and documentation
are all available at \url{https://eval-unlearn.readthedocs.io}.
% The rising number of concept unlearning techniques for text-to-image (T2I) diffusion models has
% produced a fragmented evaluation landscape. Methods are assessed under heterogeneous experimental
% conditions making principled cross-method comparison difficult. We present \evalunlearn, an open-source Python library providing a unified,
% reproducible benchmarking framework for concept unlearning in T2I Diffusion models. \evalunlearn{}
% integrates twelve published unlearning techniques spanning fine-tuning, closed-form model editing,
% and inference-time intervention, alongside nine complementary evaluation metrics covering erasure
% efficacy, adversarial robustness, generative quality, and concept retention. We demonstrate
% \evalunlearn{} through a nudity concept erasure case study across all twelve techniques, exposing
% significant accuracy-quality trade-offs that are obscured by heterogeneous evaluation via a leader-board published at huggingface (\href{https://huggingface.co/spaces/REAL-Lab-Imperial/eval-unlearn}{https://huggingface.co/spaces/REAL-Lab-Imperial/eval-unlearn}). \evalunlearn{} is
% released under the MIT license, installable via \texttt{pip~install~eval-unlearn}, with code at
% \url{https://github.com/REAL-Lab-Imperial/Eval-Unlearn} and documentation at
% \url{https://eval-unlearn.readthedocs.io}.

% The library features a Python entry-point plugin architecture enabling third-party techniques and metrics to self-register upon installation without modifying the core framework, frozen-dataclass configurations for reproducibility, streaming dataset loaders, proactive GPU memory management, and a CLI with declarative configuration files for accessibility.
\end{abstract}

\section{Introduction}
\label{sec:intro}

Text-to-image diffusion models such as Stable Diffusion~\citep{rombach2022ldm} synthesize
high-quality images from natural language prompts, but training on large, internet-scraped corpora
makes them capable of generating inappropriate content, with many negative implications
\citep{schramowski2023sld}. Fully retraining on curated data is both expensive and often
ineffective due to compositionality \citep{okawa2023compositional}. \textit{Machine unlearning}
offers a practical alternative, using targeted weight modifications or inference-time
interventions to selectively suppress a concept from a trained model
\citep{beerens2025vulnerability}.

A recent survey \citep{kim2025survey} groups T2I unlearning techniques into three categories:
\textit{fine-tuning} methods (ESD~\citep{gandikota2023esd}, CA~\citep{kumari2023ablating},
CoGFD~\citep{nie2025erasing}, AdvUnlearn~\citep{zhang2024advunlearn}, SSD~\citep{foster2023ssd})
that iteratively adjust U-Net or text-encoder weights; \textit{closed-form model editing} methods
(UCE~\citep{gandikota2024uce}, MACE~\citep{lu2024mace}) that compute single-step weight updates
without iterative optimisation; and \textit{inference-time interventions}
(SLD~\citep{schramowski2023sld}, SAFREE~\citep{yoon2025safree},
TraSCE~\citep{jain2025trascetrajectorysteeringconcept}, ConceptSteerers~\citep{kim2025conceptsteerers},
SAeUron~\citep{cywinski2025saeuron}) that reshape guidance or latent representations at inference
time without modifying weights.

Despite this taxonomy, techniques are evaluated on different datasets, metrics, and hyperparameter
regimes, making reliable cross-method comparison difficult. Existing frameworks, namely
UnlearnCanvas \citep{zhang2024unlearncanvas} and the Holistic Unlearning Benchmark
\citep{moon2025holisticunlearningbenchmarkmultifaceted}, provide fixed dataset-and-script packages
covering a narrow, largely fine-tuning/closed-form set of techniques, with no interface for
extension without modifying core code. \evalunlearn{} closes these gaps with a plugin architecture
spanning all three categories. Its contributions are: (i) a shared execution pipeline supporting
all three technique categories on a common base model; (ii) nine standardized metrics spanning
erasure efficacy, adversarial robustness, quality, and retention; (iii) a plugin architecture, with
a validation notebook for community-contributed techniques and metrics, enabling extension without
modifying the core framework; and (iv) a public HuggingFace leaderboard, logged with full
hyperparameters for reproducibility, enabling real-time comparison of unlearning techniques. The
library is further accompanied by tutorial notebooks and a test suite exercised in continuous
integration, currently at 99.82\% coverage. Figure~\ref{fig:pipeline} illustrates the resulting
pipeline.

\section{Implemented Techniques and Metrics}
\label{sec:features}

% Tables ~\ref{tab:techniques}, and ~\ref{tab:metrics} lists the twelve techniques and nine metrics
% available in \evalunlearn{} v0.1.6. All techniques target Stable Diffusion v1.4 as the base model
% (SLD couples to its safety-fine-tuned variant). The \texttt{free\_run} technique allows any custom
% checkpoint or HuggingFace model ID to be evaluated without writing wrapper code. Technique packages are separately installable so users need only the components required for their experiment. The configuration for the metric computation as well as the unlearning method is configurable from the metric\_configs and technique\_config variables respectively in the SingleBenchmarkRunner or the MultiBenchmarkRunner. The default configuration of the metrics and the methods is described in the documentation and is the accepted standardized values or the values reported in the respective techniques. 

% \input{tables/unlearning-techniques.tex}
% \input{tables/unlearning-techniques}

\evalunlearn{} v1.1.2 integrates twelve techniques spanning the three categories introduced in
Section~\ref{sec:intro} (fine-tuning, closed-form editing, inference-time intervention), all
targeting Stable Diffusion v1.4 as the base model, with the exception of SLD, which couples to its
safety-fine-tuned variant. The framework additionally allows any custom
checkpoint or HuggingFace model ID to be evaluated without writing wrapper code, and technique
packages are separately installable so users need only the components required for their
experiment.

Alongside these, the library provides nine evaluation metrics spanning four categories.
\textit{Erasure efficacy} is measured via attack success rate (ASR) on the I2P benchmark
\citep{schramowski2023sld}. \textit{Adversarial robustness} is measured via ASR under three
optimisation-based red-teaming attacks: Ring-A-Bell's genetic search \citep{tsai2024ringabell},
MMA-Diffusion's GCG suffix attack \citep{yang2023mma}, and P4D's gradient-based prompt optimisation
\citep{chin2024p4d}. \textit{Generative quality} is captured by FID and CLIP Score against COCO
2017 \citep{lin2014coco}, and by TIFA for compositional fidelity \citep{hu2023tifa}.
\textit{Concept retention} is assessed via the ERR erasure-retention metric \citep{liu2025genu} and
UA-IRA, a retention score computed over user-provided prompts.

The configuration for both the unlearning method and the metric computation is exposed through the
\texttt{technique\_config} and \texttt{metric\_configs} variables of \texttt{SingleBenchmarkRunner}
and \texttt{MultiBenchmarkRunner}, respectively. Default values follow either accepted standardised
conventions or the values reported in each metric's or technique's original publication, as
documented in the library documentation.

\section{Software Design}
\label{sec:design}

\evalunlearn{} is organised around a small set of core components following the Adapter design pattern \citep{McDonough2017AdapterDP}: the framework defines fixed abstract interfaces, while external implementations are integrated through wrapper classes, keeping orchestration logic agnostic to technique and metric internals. Technique versions are pinned to their most recent release (Sept 2026) for reproducibility.

\textbf{Technique.} A technique is a concept-erasure method exposed through a wrapper implementing a fixed \texttt{generate()} interface. Techniques are discovered via Python entry points declared in a package's \texttt{pyproject.toml}: on runner initialisation, \texttt{load\_entrypoints} scans all installed packages and registers each class under its declared name, making it immediately addressable in configuration files without manual imports or framework modification. The same mechanism applies to metrics and dataset loaders.

\textbf{Metric.} A metric computes a single evaluation score (e.g., attack success rate, fidelity) over a stream of generated images via an \texttt{update()} method called per batch and a \texttt{compute()} method called at the end of the run.

\textbf{Dataset Loader.} Dataset loaders supply prompts using one of two strategies. Algorithmically generated prompts are produced on first use and optionally cached to disk to avoid repeating costly optimisation. Fixed benchmark datasets (I2P, COCO, TIFA) are streamed from HuggingFace \texttt{datasets}, ensuring no complete dataset is held in memory.

\textbf{Configuration.} Techniques and metrics are paired with frozen dataclass configurations that validate all hyperparameters at initialisation, surfacing misconfigurations before any model weights are loaded. Each experiment writes a concise report (metric scores) and an extended report (scores plus full configuration) to an output directory identified by a unique ID derived from the technique name, metric names, configurations, and timestamp.

\textbf{Runner.} The runner ties together an unlearning technique, one or more metrics, and their configurations into a single experiment. \texttt{SingleBenchmarkRunner} executes one technique against one metric; \texttt{MultiBenchmarkRunner} evaluates one technique across multiple metrics in a single pass, reusing the loaded model. The run loop streams batches from each metric's dataset loader, calls \texttt{technique.generate(prompts)}, and accumulates statistics via \texttt{metric.update()}, with final aggregation deferred to \texttt{metric.compute()}. Metric objects are explicitly deleted and garbage-collected between evaluations to reclaim VRAM. Inference runs in FP16, with FP32 used for fine-tuning where numerical stability requires it. Runners are launched via CLI or script, driven by a JSON or YAML configuration file.

\section{Conclusion}
\label{sec:conclusion}

\evalunlearn{} provides a unified, reproducible benchmarking framework for concept unlearning
in T2I diffusion models. By standardizing the evaluation pipeline across twelve techniques and nine
metrics, it enables fair cross-method comparison that is otherwise impractical. Its plugin
architecture ensures the framework will remain current as the field evolves, and its MIT license
minimizes barriers to community adoption and contribution. Future work will support newer Stable
Diffusion variants (v2, SDXL), add computational overhead metrics (latency, peak VRAM), expand
concept coverage beyond nudity and violence.

\bibliography{references}


\end{document}